\documentclass[10pt,letterpaper]{article}
\usepackage{amssymb}  
 
\usepackage{cogsci}
\cogscifinalcopy 
\usepackage{tabularx}
\usepackage[table]{xcolor}
\usepackage{multirow}
\usepackage{booktabs}
\usepackage{hyperref}
\usepackage{graphicx}
\usepackage{pslatex}
\usepackage{apacite}
\usepackage{subcaption}
\usepackage{float} 

\hypersetup{
    colorlinks, 
    linkcolor={blue!30!black}, 
    citecolor={blue!70!black}, 
    urlcolor={blue!50!black} 
}
\definecolor{lightred}{RGB}{229, 10, 27}
\definecolor{DeepBlue}{RGB}{0, 0, 255}
\newcommand{\lightuparrow}{\textcolor{lightred}{$\mathbf{\uparrow}$}}

\newcommand{\lightdownarrow}{\textcolor{DeepBlue}{$\mathbf{\downarrow}$}}

\title{Structuralist Approach to AI Literary Criticism: Leveraging Greimas Semiotic Square for Large Language Models}
 
\author{{\large \bf Fangzhou Dong\textsuperscript{\ddag}, Yifan Zeng\textsuperscript{\ddag}, YingPeng Sang\textsuperscript{*}} \\
  School of Computer Science and Engineering, Sun Yat-sen University, Guangzhou, China \\
  \texttt{\{dongfzh@mail2, zengyf53@mail2, sangyp@mail\}.sysu.edu.cn}
  \AND{\large \bf Hong Shen} \\
  School of Engineering and Technology, Central Queensland University, Queensland, Australia\\
  \texttt{h.shen@cqu.edu.au}
  }

\begin{document}

\maketitle
\renewcommand{\thefootnote}{\fnsymbol{footnote}}
\footnotetext[3]{Co-first authors.}
\footnotetext[1]{Corresponding author.}

\begin{abstract}
Large Language Models (LLMs) excel in understanding and generating text but struggle with providing professional literary criticism for works with profound thoughts and complex narratives. This paper proposes GLASS (Greimas Literary Analysis via Semiotic Square), a structured analytical framework based on Greimas Semiotic Square (GSS), to enhance LLMs' ability to conduct in-depth literary analysis. GLASS facilitates the rapid dissection of narrative structures and deep meanings in narrative works. We propose the first dataset for GSS-based literary criticism, featuring detailed analyses of 48 works. Then we propose quantitative metrics for GSS-based literary criticism using the LLM-as-a-judge paradigm. Our framework's results, compared with expert criticism across multiple works and LLMs, show high performance. Finally, we applied GLASS to 39 classic works, producing original and high-quality analyses that address existing research gaps. This research provides an AI-based tool for literary research and education, offering insights into the cognitive mechanisms underlying literary engagement.

\textbf{Keywords:} 
Large Language Models; Literary Criticism; Greimas Semiotic Square
\end{abstract}

\section{Introduction}
Literary criticism \cite{LiteraryCritism}, guided by literary theory, is a scientific and objective interpretative activity that analyzes and evaluates literary works. Unlike subjective criticism, it creatively explores the artistic, emotional, and ideological content, narrative patterns, and deep meaning structures of texts. Rooted in theories like Gadamer’s fusion of horizons \cite{Gadamer}, it merges new perspectives with the original work’s context. Its significance lies in offering readers deeper aesthetic insights and understanding of the ideologies embedded in the work, thereby enhancing their criticism. \textbf{As a cognitive activity} \cite{CogLiterary,CogLiter2,CogLiterary1,CogLiter3}, literary criticism explores how readers process and integrate textual information. It involves parsing symbols, metaphors, constructing networks of meaning, predicting narrative developments, and engaging in critical reflection. This cognitive dimension highlights its role in understanding the mechanisms of human comprehension and engagement with literature.

LLMs \cite{LLMCog1} have shown impressive capabilities in natural language understanding and generation, with natural language serving as the medium for narrative patterns and meaning structures in literary works. Therefore, utilizing LLMs for in-depth assistance in literary research is a new research paradigm \cite{LLMgenpoem,GPT4Meto}, integrating computational thinking with traditional qualitative approaches \cite{NewLiterary}. This will provide the field of literary studies with novel and diverse perspectives and inspirations, while also assisting beginners in learning literary. This paper will explore a method for LLMs to conduct relatively profound and professional literary criticism that embody human art, emotions, thoughts, complex narrative patterns, and deep meaning structures.

Furthermore, from the interdisciplinary perspective of cognitive science, by viewing literary works as complex symbolic systems, LLM as an AI reader can offer new insights into the cognitive processes. It can reflect the AI reader's cognitive structure because literary reading is a cognitive activity that involves understanding, interpreting texts, and integrating personal experiences. Moreover, within the framework of cognitive science, Greimas Structuralist Semiotic Square can be employed to analyze cognitive structures and narrative comprehension, as people often organize the world through binary oppositions. By integrating LLMs with Greimas Semiotic Square, we not only develop a powerful LLM-based literary criticism tool, but also enhance our understanding of how AI systems, as readers, process meaning and narrative structures, thereby helping shed light on the structured cognitive mechanisms underlying literary engagement.

However, compared to professional human literary scholars, the literary criticism, narrative structure analysis, and meaning interpretation provided by LLMs are often superficial, uninteresting, and overly general \cite{LLMjuxian}. Firstly, LLMs often lack the ability to apply literary theories (such as structuralism, feminism, post-colonialism, psychoanalysis, existentialism, and Marxism) \cite{Structuralism,Female,LiteraryTheroy,Marxism} to provide in-depth and systematic criticism. Secondly, LLMs may fail to fully comprehend metaphors, allusions, and socio-cultural contexts within specific cultural backgrounds. Thirdly, LLMs, which are primarily based on pre-trained data, struggle to generate truly original and high-quality insights. Lastly, LLMs cannot flexibly integrate interdisciplinary knowledge (such as philosophy, psychology, and history) to analyze literary works, as human scholars do. Moreover, current evaluation of generated literary criticism relies on subjective judgment by scholars, leading to potential biases, inconsistent opinions, and low efficiency. Obtaining high-quality scholarly assessments is costly and challenging. A possible solution is developing automated, standardized evaluation tools.

\textbf{Contributions.} To address these challenges and investigate how AI can simulate the cognitive processes of readers using structuralist methods to analyze literary works, we apply the Greimas semiotic square \cite{GremiasOnMeaning,GremiasStructure}. This provides LLMs with a structured analytical framework called \textbf{GLASS} (\textbf{G}reimas' \textbf{L}iterary \textbf{A}nalysis via \textbf{S}emiotic \textbf{S}quare). This framework helps overcome inherent limitations in LLMs' literary criticism capabilities and serves as a model for demonstrating how AI can simulate reader cognition using structuralist analysis. Our contributions are:
\vspace{-3pt}
\begin{itemize}
    \item We propose GLASS for LLM-based literary criticism and narrative meaning structure analysis. It can achieve rapid dissection of narrative structures and deep meanings. To our knowledge, we are the first to utilize Greimas semiotic square with LLMs for analyzing narrative works.

    \item To our knowledge, we are the first to construct and propose a dataset for Greimas Semiotic Square-based literary criticism, aimed at training and evaluating LLMs on this task. The dataset includes detailed analyses of 49 narrative works (39 high-quality entries generated by our framework and 10 from professional scholars' papers), covering 4 items, 6 relations and a summary for each work.

    \item We applied GLASS to conduct novel and in-depth structuralist semiotic square analyses on 39 narrative works, showcasing how AI can mimic the cognitive processes involved in literary interpretation.

    \item We propose quantitative metrics for GSS-based literary criticism using the LLM-as-a-judge paradigm. Our framework's results are extensively evaluated and compared with expert criticism across multiple works and LLMs. The findings demonstrate that our framework outperforms professional literary scholars in accuracy, completeness, logic, and inspiration.

\end{itemize}

\section{Proposed Framework GLASS}

\subsection{About Greimas semiotic square}
Structuralist literary criticism theory originated from many outstanding scholars, such as Roland Barthes \cite{Barthes}, Claude Lévi-Strauss \cite{LeviStrauss}, and also includes A. J. Greimas \cite{AJGreimas}. Greimas was inspired by Propp's study of characters in folktales, Lévi-Strauss's binary oppositions, Aristotle's logical propositions (contradictory and contrary), and linguists Saussure and Jakobson (meaning generated through opposition of semes). Building on the foundation of binary opposition, Greimas expanded this concept into a four-term system, proposing the semiotic square. This mode of semiotic analysis takes meaning as its starting point, using semantic relationships to deduce the deep structure hidden beneath the surface structure, and establishes a relatively complete narrative theory. Greimas believed that meaning can only be generated through the opposition of semes \cite{GremiasOnMeaning}.

Specifically, the semiotic square analysis process is as follows: Establish a term $X$, its opposite is anti-$X$. Beyond this, there is non-$X$, which contradicts $X$ but is not necessarily opposite to it, and there is also non-anti-$X$, which contradicts anti-$X$. $X$ and anti-$X$ form a core binary opposition, representing a relationship of mutual absolute negation. non-$X$ and non-anti-$X$ constitute a secondary opposition. non-$X$ contradicts $X$ but is not necessarily opposite to it, and the same applies to non-anti-$X$ and anti-$X$.
In Greimas's view, stories originate from the core opposition between $X$ and anti-$X$, but as the story progresses, numerous new logical elements are derived, leading to non-$X$ and non-anti-$X$. When all these aspects are fully developed, the story is complete.
It is through the relationships between these four functional terms in the diagram and their various transformations that the meaning of the story is derived. 

Greimas' Semiotic Square is significant not only in literary criticism and semiotics but also in its intersection with cognitive science. The Semiotic Square emphasizes that meaning emerges through opposition and contradiction, echoing cognitive science theories. Humans often rely on cognitive strategies such as contrast, opposition, and categorization to understand and construct meaning in complex environments. Additionally, the Semiotic Square provides a formal model for analysis, which is useful for modeling human cognition and understanding the mind.

In the field of AI and natural language processing (NLP), Greimas' Semiotic Square can help machines understand the deep structure and semantic relationships in texts. Applying it to LLMs enhances their text comprehension and semantic analysis capabilities. This application reflects the close connection between cognitive science and AI, both of which explore how to better understand and generate language.

\begin{figure}[t]
  \centering
    \includegraphics[width=1\columnwidth]{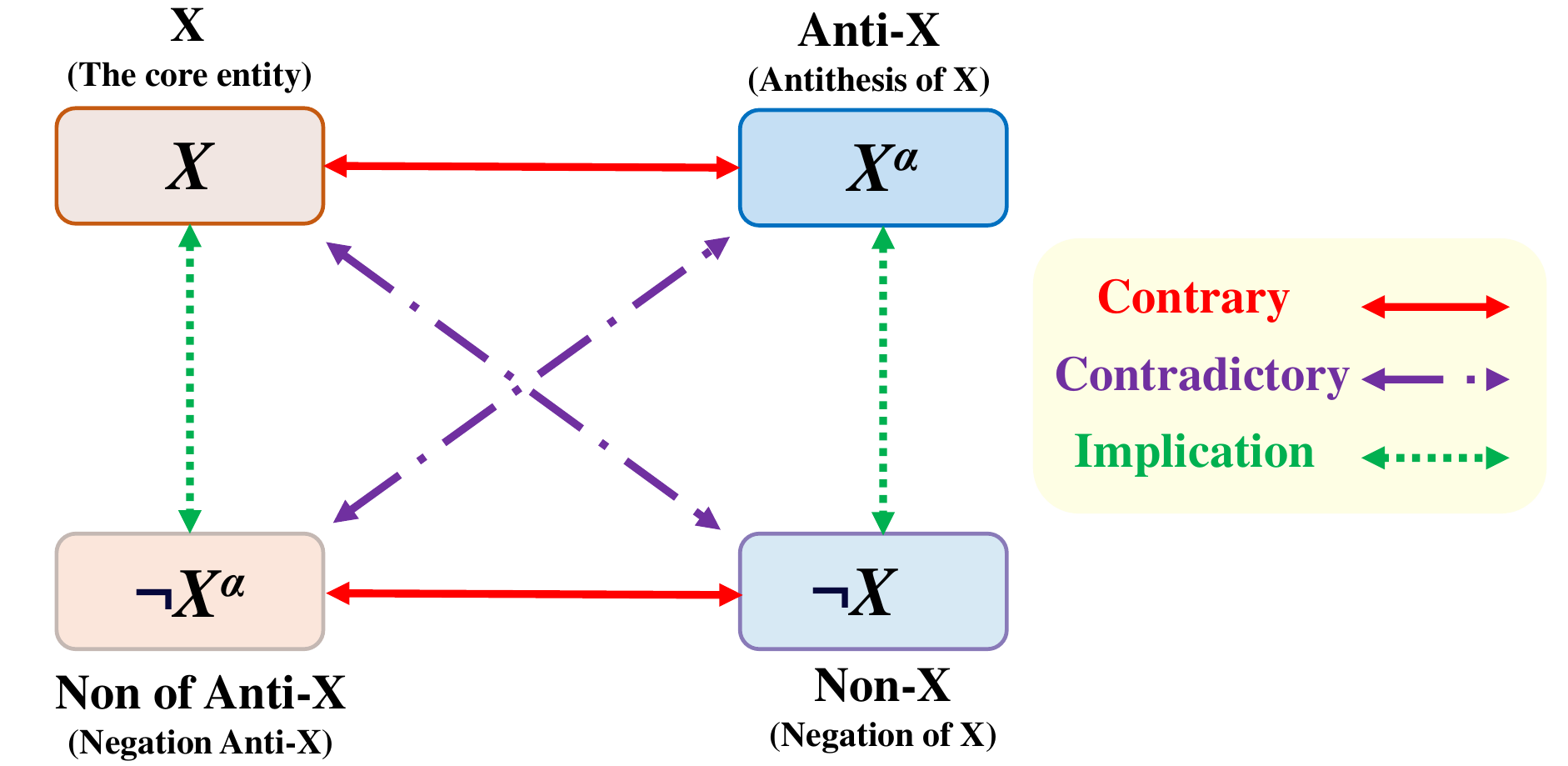}
    \label{fig:sub2}
  \vspace{-10pt}
  \caption{Example of Greimas semiotic square.}
  \label{fig:test}
\end{figure}

\begin{figure*}[h]
  \includegraphics[width=\linewidth]{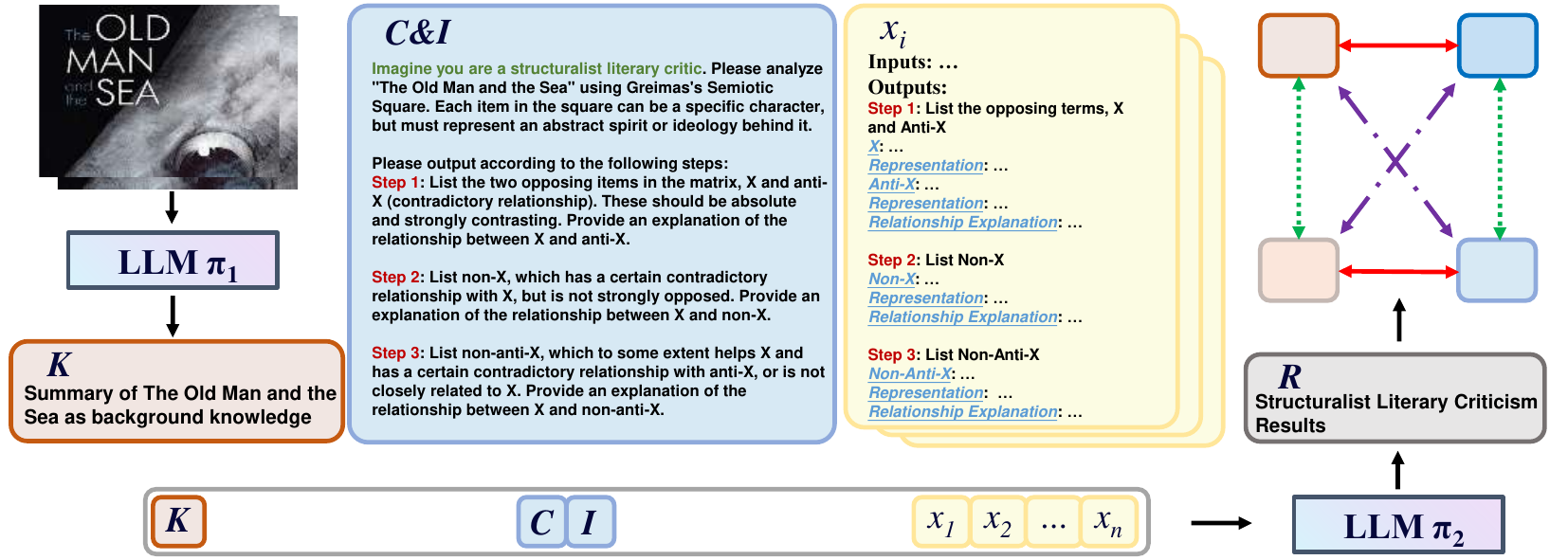}
  \caption{The GLASS framework diagram illustrates LLM-based literary work analysis of Hemingway's \textit{The Old Man and the Sea} using Greimas Semiotic Square. By fusing various prompt techniques and analyzing samples with multiple semiotic squares via few-shot prompting, 4 logical factors and 6 relationships are derived, thereby unraveling the profound meaning of the story. LLM $\pi_1$ and LLM $\pi_2$ can be different LLMs to avoid contextual interference.}
  \label{fig:framework}
\end{figure*}

\begin{figure*}[h]
  \centering
  \subcaptionbox{}{\includegraphics[width=0.3\linewidth]{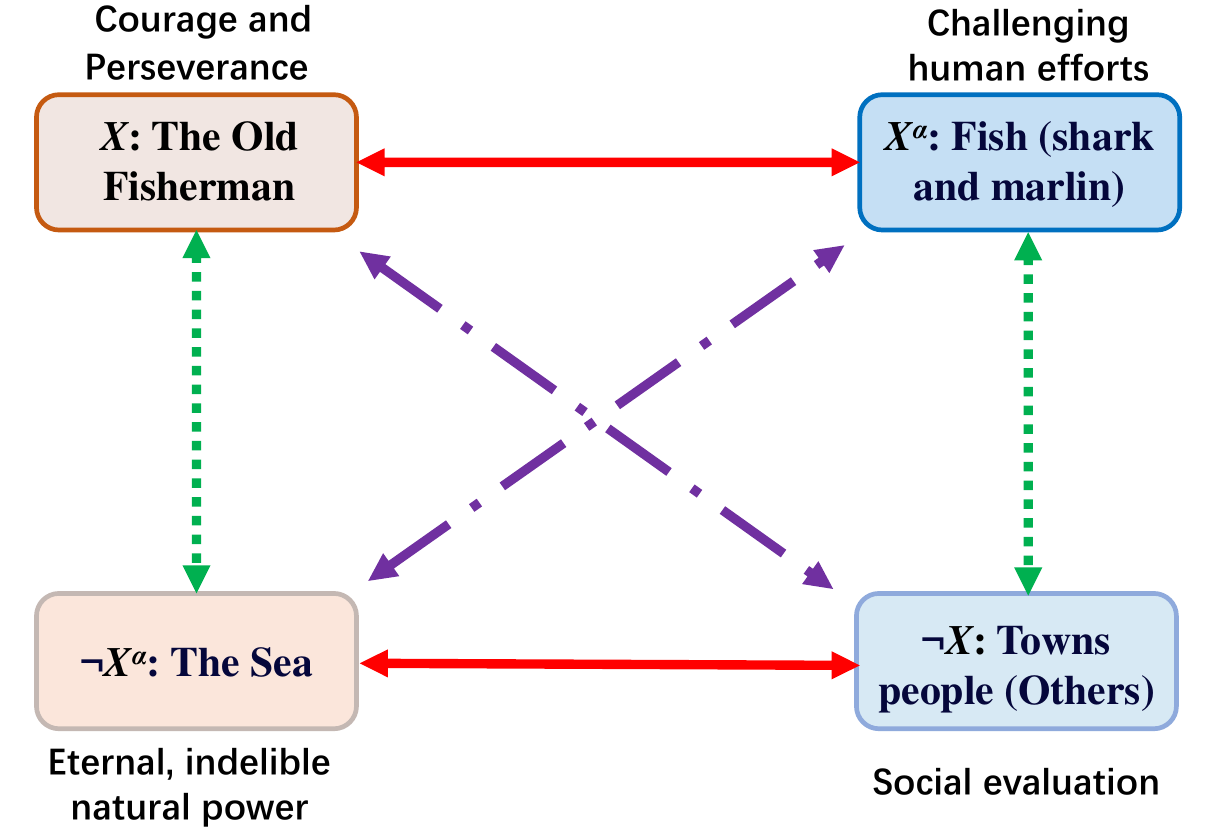}}
  \hfill
  \subcaptionbox{}{\includegraphics[width=0.3\linewidth]{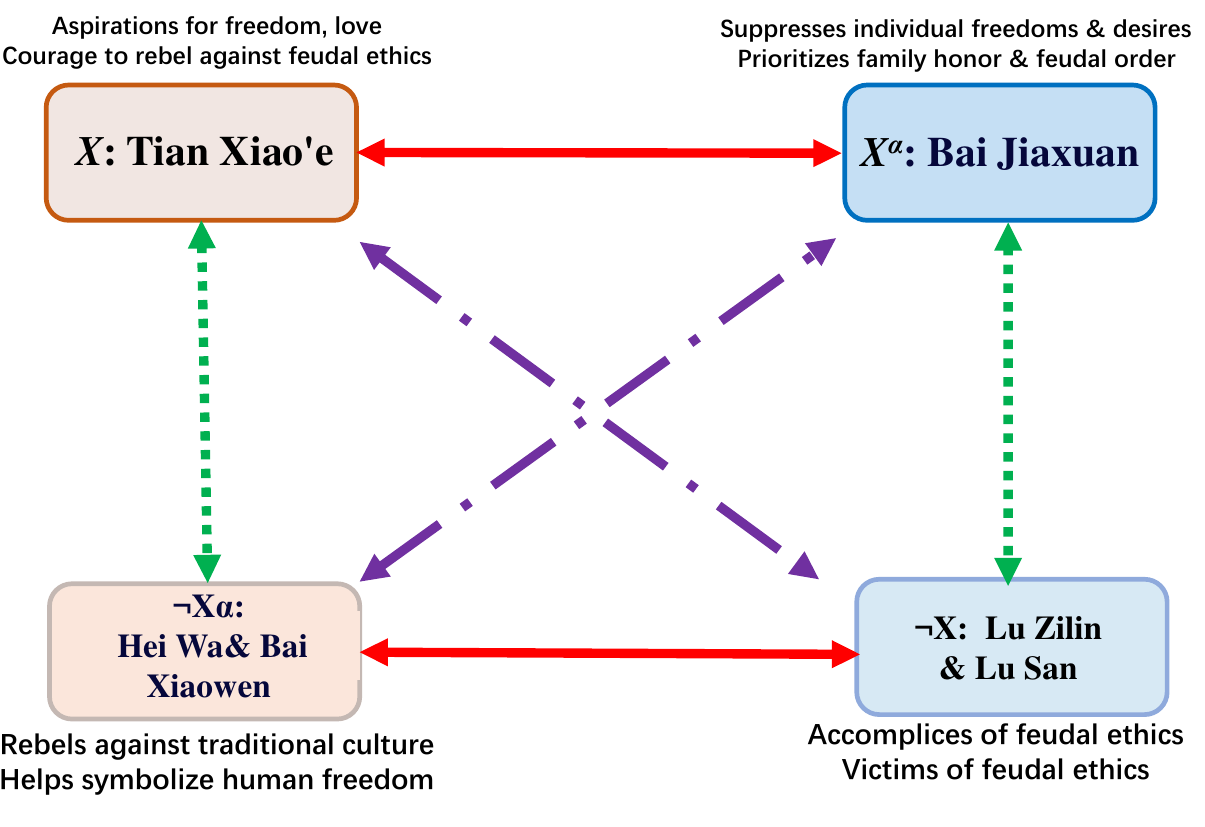}}
  \hfill
  \subcaptionbox{}{\includegraphics[width=0.3\linewidth]{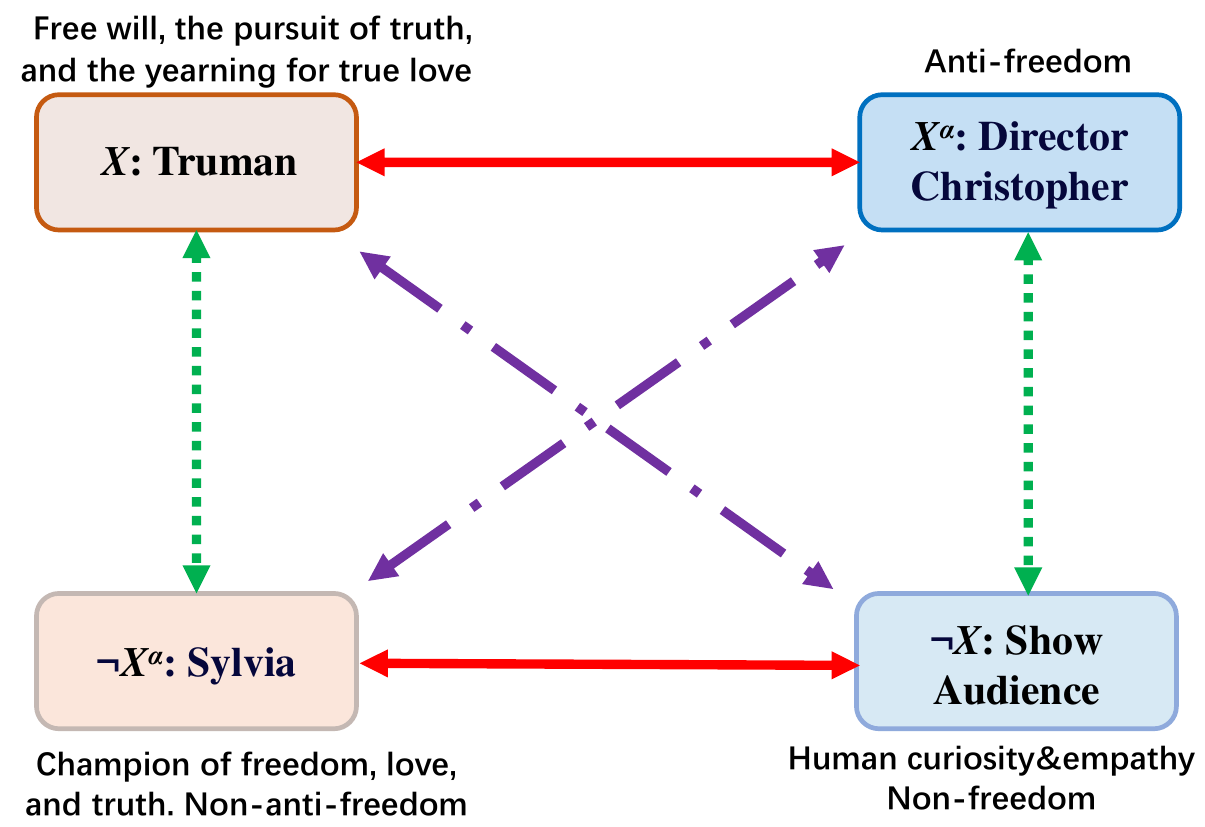}}
  \hfill
  \subcaptionbox{}{\includegraphics[width=0.3\linewidth]{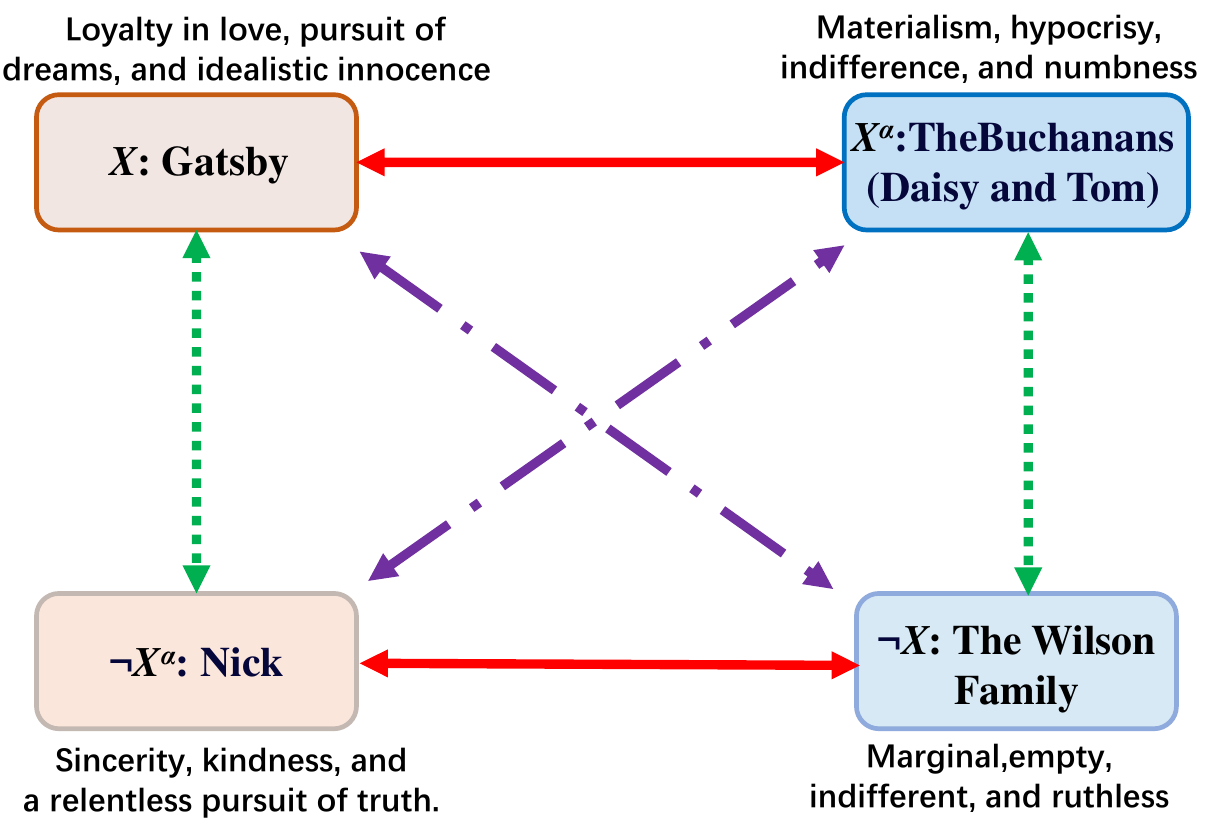}}
  \hfill
  \subcaptionbox{}{\includegraphics[width=0.3\linewidth]{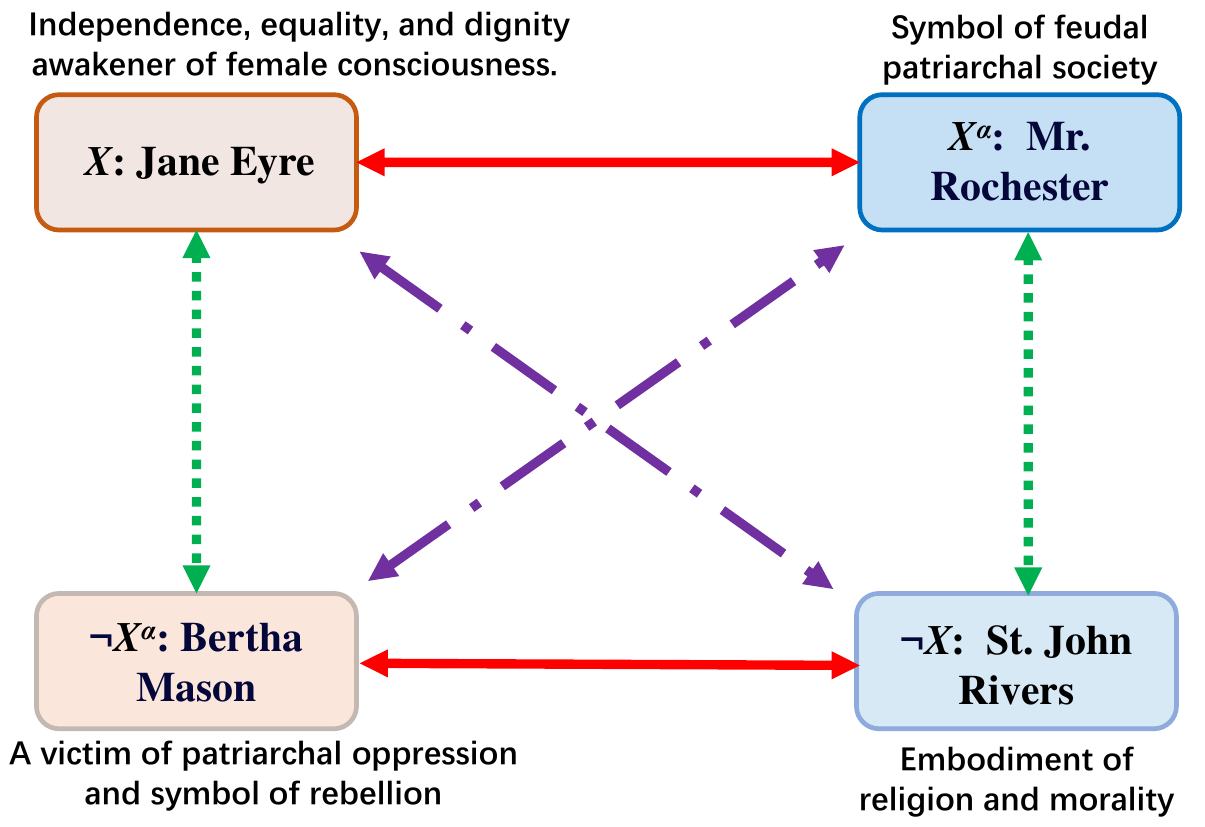}}
  \hfill
  \subcaptionbox{}{\includegraphics[width=0.3\linewidth]{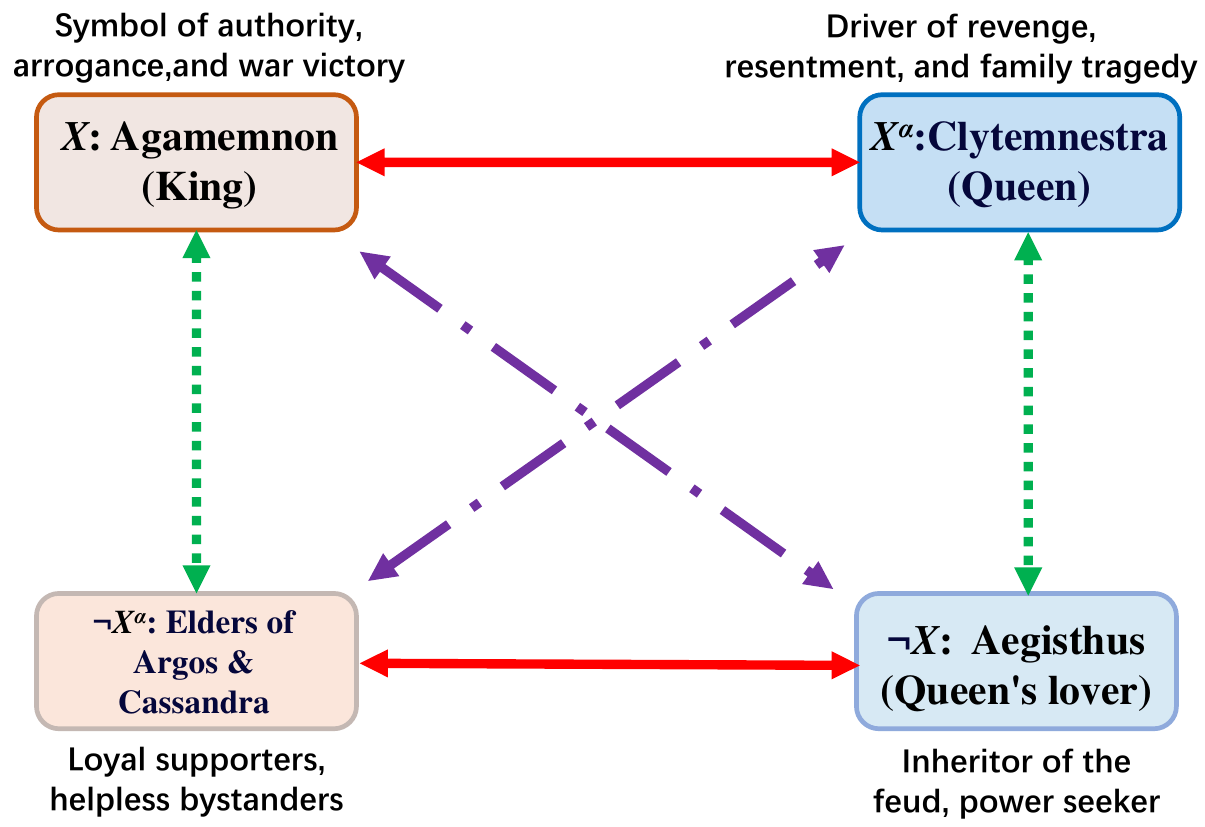}}
  \vspace{-10pt}
  \caption{Greimas semiotic square of part of our proposed dataset. (a) \textit{The Old Man and the Sea}. (b) \textit{White Deer Plain}. (c) \textit{The Truman Show}. (d) \textit{The Great Gatsby}. (e) \textit{Jane Eyre}. (f) \textit{Agamemnon}.}
  \label{fig:GSSExp}
\end{figure*}


\subsection{Semiotic square for LLMs}
GLASS enhances LLMs' reasoning, improves coherence, and uses LLMs to automatically generate these squares as "semantic probes." GLASS extracts core semantic axes (e.g., "life - death" and "action - hesitation" in \textit{Hamlet}), facilitating literary analysis by boosting interpretative capabilities, generating opposing concepts  and multidimensional insights.

\textbf{Systematicness and standardization.} GLASS utilizing semiotic square to offer a systematic, standardized, and scientific method for LLMs to analyze binary contrary, contradictory and implication in texts. By focusing on objective logical relationships, it reduces dependence on cultural and historical background knowledge, minimizing LLM errors due to knowledge limitations or biases.

 \textbf{Effectiveness and profundity.} GLASS leveraging semiotic square analysis, which reveals binary oppositions between characters and ideologies, as well as complex implications and contradictions, bringing the text's deep structure and implicit meanings to the surface (making them easier for LLMs to identify). This structured thinking process helps LLMs produce profound insights. The semiotic square generated by LLMs clearly reveals the narrative's logic, as well as implicit ideologies, through opposition and contradiction.

\textbf{Prompt techniques.} The prompt structure for GLASS follows a specific formula (1). It begins with the generation of a literary summary $K$ by LLM $\pi_1$ based on its vast pre-training data and the knowledge acquired during training. $K$ serves as a knowledge background to assist the LLM $\pi_2$. Experiments have proven that $K$ enhances the capabilities of LLM across datasets \cite{GKP}. Next, there is the context $C$, which specifies the role the LLM $\pi_2$ needs to assume: a literary expert. Following this is a clear instruction $I$, employing a chain-of-thought (CoT) \cite{CoT} approach to guide the model in constructing the sign matrix and filling in the semantic relationships step by step. Finally, there is  few-shot prompting, where numerous input-output examples $\{x_1,x_2,...,x_n\}$ are formatted and provided, allowing us to populate the dataset.
\vspace{-3pt}
\begin{equation}
    \mathbf{Prompt}_i = [K_i, C_i, I_i, x_1, x_2,..., x_n]
\end{equation}

\section{Proposed Dataset}

\begin{figure}[h]
  \centering
  \includegraphics[width=1\columnwidth]{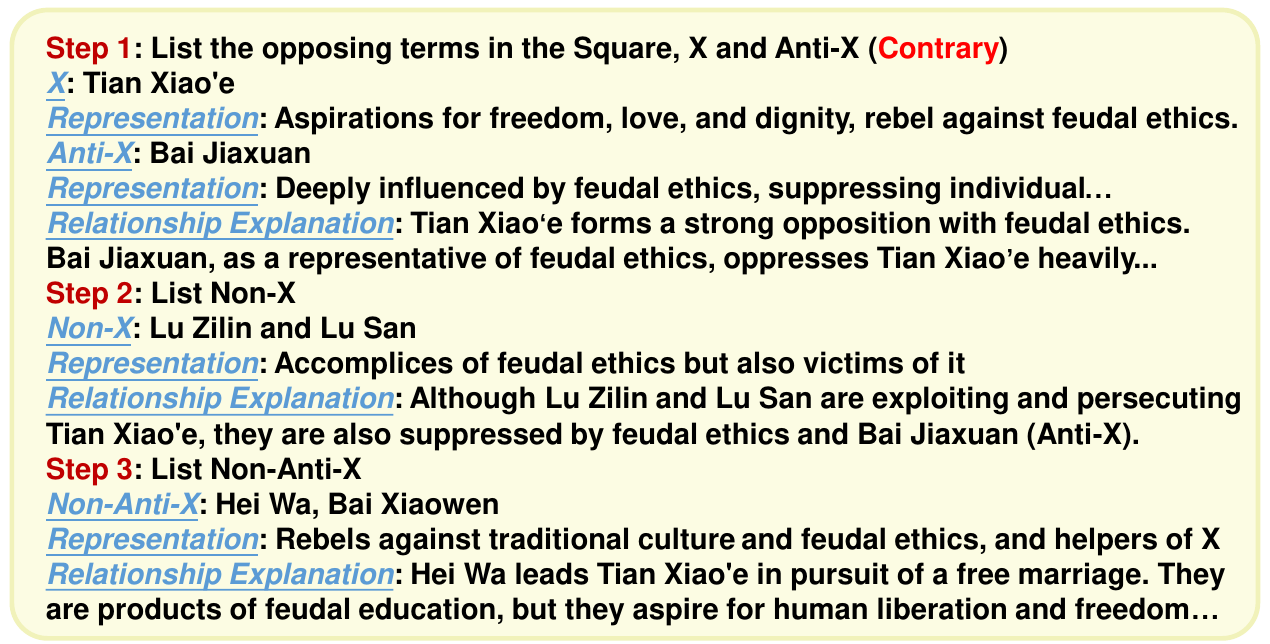}
  \vspace{-10pt}
  \caption{A sample of \textit{White Deer Plain} in our proposed dataset. Using the (Key:Value) prompt format, along with CoT, can improve the performance of LLMs.}
  \label{fig:sampleData}
\end{figure}


We constructed a Greimas semiotic square-based dataset for literary criticism, aimed at training and evaluating LLMs in structuralist interpretation. The dataset includes detailed analyses of 49 narrative works (39 by our framework, 10 from scholars), covering 4 items, 6 relations, and a summary per work. Selected texts are novels or film narratives with significant depth, representing diverse genres and cultural backgrounds. Our dataset comprises expert literary criticisms and LLM-generated analyses. We surveyed and read papers (cited in next paragraph) by literary scholars using the Greimas semiotic square for literary criticism, collected and formatted analysis examples.

The first part of the dataset includes analyses of works such as \textit{The Old Man and the Sea} by Hemingway \cite{OldMan}, \textit{White Deer Plain} by Chen Zhongshi \cite{White}, \textit{Christmas Eve} by Gogol \cite{Christmas}, the film \textit{The Truman Show} \cite{Trueman}, \textit{Jane Eyre} by Charlotte Brontë \cite{Jane}, the film \textit{Spirited Away} by Hayao Miyazaki \cite{Spirited}, \textit{The Great Gatsby} by Fitzgerald \cite{Great}, \textit{Pride and Prejudice} by Jane Austen \cite{Pride}, and the tragedy \textit{Agamemnon} by Aeschylus \cite{Agamemnon}.

The second part of the dataset consists of content generated by LLMs using the framework. The LLMs employed include multiple current state-of-the-art proprietary models:  \includegraphics[height=0.7em]{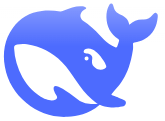} \textbf{DeepSeek-R1}\footnote{\url{https://chat.DeepSeek.com/}}, \includegraphics[height=0.7em]{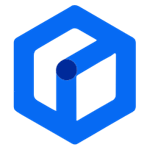} \textbf{ERNIE 3.5}\footnote{\url{https://yiyan.baidu.com/}}, \includegraphics[height=0.7em]{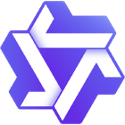} \textbf{Qwen2.5} \footnote{\url{https://tongyi.aliyun.com/}}, \includegraphics[height=0.6em]{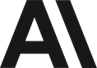} \textbf{Claude 3.5 Sonnet}\footnote{\url{https://claude.ai/new}}, and \includegraphics[height=0.7em]{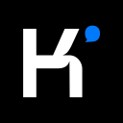} \textbf{Kimi}\footnote{\url{https://kimi.moonshot.cn/}}. The works include \textit{One Hundred Years of Solitude}, \textit{War and Peace}, \textit{The Hunchback of Notre-Dame}, \textit{Robinson Crusoe}, \textit{The Little Prince}, \textit{Hamlet}, \textit{Don Quixote}, \textit{Gone with the Wind}, \textit{The Catcher in the Rye}, \textit{A Brief History of Time}, \textit{Crime and Punishment}, \textit{Wuthering Heights}, \textit{A Tale of Two Cities}, \textit{Oliver Twist}, \textit{Kafka on the Shore}, \textit{The Shawshank Redemption}, \textit{Forrest Gump}, \textit{The Kite Runner}, \textit{To Kill a Mockingbird}, \textit{Little Women}, \textit{The Moon and Sixpence}, \textit{Walden}, \textit{Titanic}, \textit{Schindler's List}, \textit{Inception}, \textit{The Godfather}, \textit{The Matrix}, \textit{Interstellar}, \textit{Zootopia}, \textit{Bohemian Rhapsody}, \textit{Game of Thrones}, and \textit{The Grand Budapest Hotel}.

The dataset primarily consists of classic works that hold significant positions in literary and film history. The selected works cover diverse cultural and historical backgrounds, including British, American, Ancient Greek, Japanese, and Chinese works from various periods (Ancient Greece, Victorian era, 20th century, etc.). This diversity provides LLMs with rich learning material on different literary modes across various cultural, historical, and spatiotemporal contexts.

Each sample in the dataset includes a Greimas semiotic square, explanations of the oppositions between characters and their represented abstract spirits and values (i.e., the relationships between various semes) as illustrated in Figure~\ref{fig:GSSExp} and \ref{fig:sampleData}. The dataset is presented in CoT \cite{CoT} format, can integrate into LLM prompts for few-shot learning \cite{few-shot1,few-shot2} as illustrated in Figure~\ref{fig:framework}. Research suggests that integrating a few samples into prompts can perform well and even outperform fine-tuning with more samples \cite{FewshotPrompt}.

\section{Case Study: Demonstrating and Validating GLASS Framework with \textit{Journey to the West}}
To illustrate our framework's process and effectiveness, we apply the GLASS framework to analyze the classic Chinese novel \textit{Journey to the West}. We then compare the LLM's output using this framework with authoritative critiques to validate its depth and validity in literary analysis.

\textit{Journey to the West} is a classic Chinese novel by Wu Cheng'en, depicting Tang Sanzang and his three disciples' pilgrimage to obtain Buddhist scriptures, overcoming eighty-one tribulations and various demons.

\textbf{Authoritative literary criticism of \textit{Journey to the West.}} 
 Xie Zhaoyi\cite{xie1600} posits that \textit{Journey to the West} explores the theme of seeking reassurance through Sun Wukong's journey from indulgence to taming. His transformation from rebellion to reverence for Buddhist teachings symbolizes a shift from indulgence to restraint in human nature. Professor Zhu Hongbo\cite{zhu2022} asserts that \textit{Journey to the West}'s classic status is attributed to its broad readership and unique mythological elements. With the central theme of retrieving Buddhist scriptures, it integrates religious, mythological, and humanistic dimensions. Sa Mengwu\cite{sa1981} examines \textit{Journey to the West} through a political lens, viewing it as a metaphor for ancient Chinese politics. He notes that the characters and plots reflect the officialdom and social conditions of the Ming Dynasty.






\textbf{Criticizing \textit{Journey to the West} through the GLASS framework.}
Based on the GLASS framework proposed earlier for criticizing \textit{Journey to the West}, the LLMs provides the following analysis:

\begin{itemize}
    \item \textbf{Firstly}, the two opposing items in the matrix, $X$ and anti-$X$ (contradictory relationship) are listed. $X$ here denotes idealism and quest for enlightenment. Tang Sanzang's journey symbolizes the pursuit of knowledge and spiritual growth, overcoming obstacles to achieve enlightenment. anti-$X$ here denotes materialism and rebellion against authority. Characters like Sun Wukong embody materialism and rebellion, rejecting spiritual constraints, as seen in his rebellion against heavenly authority.

    \item \textbf{Secondly}, non-$X$, which has a certain contradictory relationship with $X$, but is not strongly opposed, is listed. non-$X$ here denotes skepticism and practical wisdom. Characters like Zhu Bajie and Sha Wujing show skepticism and practical wisdom, balancing the idealism of the quest and reminding readers of realworld challenges.

    \item \textbf{Finally}, non-anti-$X$, which aids $X$ and has a contradictory relationship with anti-$X$ is listed. non-anti-$X$ here denotes compliance and submission to authority. Characters like heavenly gods and Buddha embody this spirit, which is crucial for the mission's success. Sun Wukong's eventual submission to Buddha also reflects this.

    \item \textbf{Summary of Relationships:} $X$ (idealism and quest for enlightenment) vs. anti-$X$ (materialism and rebellion) highlights the core conflict. non-$X$ (skepticism and practical wisdom) balances idealism, while non-anti-$X$ (compliance and submission) supports the spiritual journey.

    \item \textbf{Conclusion:} analyzing \textit{Journey to the West} using Greimas Semiotic Square reveals the complex interplay between idealism, rebellion, skepticism, and compliance, shaping the novel's narrative and thematic structure.
\end{itemize}
\textbf{Comparative analysis.} By comparing the criticism results generated by this framework with those of authoritative scholars, we can see that our framework has touched upon several perspectives already raised by the authorities, such as religion, rebellion and submission, and political wisdom. Moreover, it has conducted an in-depth analysis from the perspective of idealism. This shows that, compared to the superficial and parallel multi-threaded analyses directly generated by LLMs without the aid of this framework, LLMs supported by this framework can provide well-reasoned and in-depth analyses on a particular theme while maintaining a broad coverage. This effectively addresses the current issues in the field of literary criticism where LLMs tend to produce either baseless or superficial analyses.

\begin{table}[t]
\footnotesize
\centering
\begin{tabularx}{\columnwidth}{Xc} 
\toprule
\textbf{Dimension} & \textbf{Weight} \\ 
\midrule
\footnotesize Core Binary Opposition Identification and Accuracy  & 25 \\ 
Extension of Oppositional Relationships & 25 \\ 
Completeness and Logicality of the Semiotic Square & 20 \\ 
Integration of Textual Details & 15 \\ 
Innovation and Inspirational Value & 15 \\ 
\textbf{Total Score} & \textbf{100} \\ 
\bottomrule
\end{tabularx}
\caption{\footnotesize{Scoring criteria and corresponding weights for QEMG using the LLM-as-a-judge paradigm.}}
\label{tab:score}
\end{table}

\begin{table*}[t]
\fontsize{7.8}{9}\selectfont
\centering
\begin{tabular}{lcccccccccccccccccccc}
\toprule
\multirow{1}{*}{\centering \textbf{{Models}}} & \multicolumn{2}{c}{\footnotesize \textbf{OMS}} & 
\multicolumn{2}{c}{\footnotesize \textbf{WDP}} &
\multicolumn{2}{c}{\footnotesize \textbf{TNBC}} & 
\multicolumn{2}{c}{\footnotesize \textbf{TTS}} 
 & \multicolumn{2}{c}{\footnotesize \textbf{JE}} & 
\multicolumn{2}{c}{\footnotesize \textbf{SA}} &
\multicolumn{2}{c}{\footnotesize \textbf{TGG}} & 
\multicolumn{2}{c}{\footnotesize \textbf{PAP}} & 
\multicolumn{2}{c}{\footnotesize \textbf{AGA}} &
\multicolumn{2}{c}{\footnotesize \textbf{MET}}\\
\cmidrule(lr){2-21} 
&\textbf{$\mathcal{L_{\textit{c}}}$} & \textbf{$\mathcal{O}_{\textit{f}}$}  &
\textbf{$\mathcal{L_{\textit{c}}}$} & \textbf{$\mathcal{O}_{\textit{f}}$}  &
\textbf{$\mathcal{L_{\textit{c}}}$} & \textbf{$\mathcal{O}_{\textit{f}}$}  &
\textbf{$\mathcal{L_{\textit{c}}}$} & \textbf{$\mathcal{O}_{\textit{f}}$}  &
\textbf{$\mathcal{L_{\textit{c}}}$} & \textbf{$\mathcal{O}_{\textit{f}}$}  &
\textbf{$\mathcal{L_{\textit{c}}}$} & \textbf{$\mathcal{O}_{\textit{f}}$}  &
\textbf{$\mathcal{L_{\textit{c}}}$} & \textbf{$\mathcal{O}_{\textit{f}}$}  &
\textbf{$\mathcal{L_{\textit{c}}}$} & \textbf{$\mathcal{O}_{\textit{f}}$}  &
\textbf{$\mathcal{L_{\textit{c}}}$} & \textbf{$\mathcal{O}_{\textit{f}}$}  &
\textbf{$\mathcal{L_{\textit{c}}}$} & \textbf{$\mathcal{O}_{\textit{f}}$} 

\\
\midrule
\includegraphics[height=1em]{Kimi.jpg} \textbf{Kimi} & 85  & \lightuparrow94 &  91  &  91 & 87  &\lightuparrow 88  & 90 & \lightuparrow91 &91 &\lightuparrow92 &86 & \lightuparrow90 & 88 &\lightuparrow89 & 90 &\lightdownarrow83 &90 &\lightuparrow94&90&90\\
\includegraphics[height=1em]{Qwen.png} \textbf{Qwen2.5} &90 &\lightuparrow94 &90 & \lightuparrow94 &83 &\lightuparrow93 &92 &\lightuparrow94 &90&\lightuparrow92&93&93 &83&\lightuparrow92 &91 &\lightuparrow94&90 &\lightuparrow94 &94&94\\
\includegraphics[height=1em]{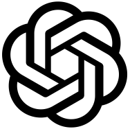} \textbf{GPT-4o mini} &86&\lightuparrow88 &87&87 &92&\lightuparrow93 &89 &\lightuparrow91 &92&\lightdownarrow87 &90&\lightdownarrow88 &89&\lightuparrow90 &86&\lightuparrow89 &86 &\lightuparrow91&93&\lightuparrow96\\ 
\includegraphics[height=1em]{OpenAI.png} \textbf{GPT-4o} &94&\lightuparrow96& 92&\lightdownarrow91 & 91&\lightuparrow92 &93&\lightuparrow95 &93&\lightdownarrow91& 91&\lightuparrow93 &93&\lightuparrow94 &91&91 &91&\lightuparrow92&92&\lightuparrow95\\ 

\bottomrule
\end{tabular}

\caption{$\mathcal{O}_{\textit{f}}$ is the score from our framework, while $\mathcal{L_{\textit{c}}}$ is the score from literary scholars using smiotic square analysis. A red arrow indicates our framework scored higher. \textbf{OMS}: \textit{The Old Man and the Sea}, \textbf{WDP}: \textit{White Deer Plain},
\textbf{TNBC}: \textit{The Night Before Christmas},
\textbf{TTS}: \textit{The Truman Show},
\textbf{JE}: \textit{Jane Eyre},
\textbf{SA}: \textit{Spirited Away},
\textbf{TGG}: \textit{The Great Gatsby},
\textbf{PAP}: \textit{Pride and Prejudice},
\textbf{AGA}: \textit{Agamemnon},
\textbf{MET}: \textit{Metamorphosis}.}
\label{tab:DomainsOfLLMs}
\end{table*}

\section{Quantitative Evaluation of GLASS}
In the case study of \textit{Journey to the West}, comparing framework-generated analysis with authoritative scholars' views showcased its effectiveness. However, subjective assessments by human scholars are variable, time-intensive, and lack comparability. In contrast, our proposed Quantitative Evaluation Metrics for Greimas semiotic square analysis (\textbf{QEMG}), using standardized metrics, ensures objectivity, comparability, and reproducibility, significantly improving evaluation efficiency for large-scale testing.
Using the LLM-as-a-judge paradigm \cite{JudgeAs}, QEMG enables various LLMs to evaluate and score outputs from GLASS and human expert comments, followed by a comparative analysis of their performance metrics.

The QEMG (in Table \ref{tab:score}) systematically evaluates GSS-based literary analyses through 5 key dimensions:  
\vspace{0.1em}  
\noindent  
\textbf{Core Binary Opposition Identification (0--25):} Accuracy in identifying core contradictions.  
\textbf{Oppositional Relationship Extension (0--25):} Depth in expanding/reconciling binary oppositions.  
\textbf{Semiotic Square Completeness (0--20):} Logical coherence and structural integrity.  
\textbf{Textual Detail Integration (0--15):} Evidence-based alignment with textual specifics.  
\textbf{Innovation and Inspiration (0--15):} Novelty and interpretative depth. QEMG emphasize \textbf{standardized metrics} (preset scoring ranges, e.g., 20--25), integration of \textbf{Greimasian logic} with quantifiable criteria, \textbf{scalability} across texts and LLM-driven validation, and \textbf{reproducibility} through predefined rubrics to minimize bias. Detailed scoring guidelines and prompts are openly accessible.

\subsection{Results and Analysis}
We employed four state-of-the-art proprietary LLMs (\textbf{Kimi}, \textbf{Qwen2.5}, \textbf{GPT-4o mini}, \textbf{GPT-4o}) as evaluators to compare the content generated by our framework against 10 professional insights extracted from human literary scholars' published papers (cited in section 3), across ten classic narrative works, using the OEMG evaluation metrics. The experimental results are presented in Table 2. Overall, the content generated by our framework demonstrated significantly higher quality compared to human literary criticism. Specifically, out of the 40 comparisons conducted across the 10 works and 4 LLMs, our framework's output was superior or on par with human experts in 34 instances (\textbf{85\%} of cases), with 29 instances (\textbf{72.5\%}) showing higher scores than human experts. These quantitative metrics clearly illustrate that the literary criticism content produced by our proposed framework is of high quality, excelling in multiple aspects.

\begin{figure}[H]
  \centering
  \subcaptionbox{\textit{Blade Runner 2049}}{\includegraphics[width=1\columnwidth]{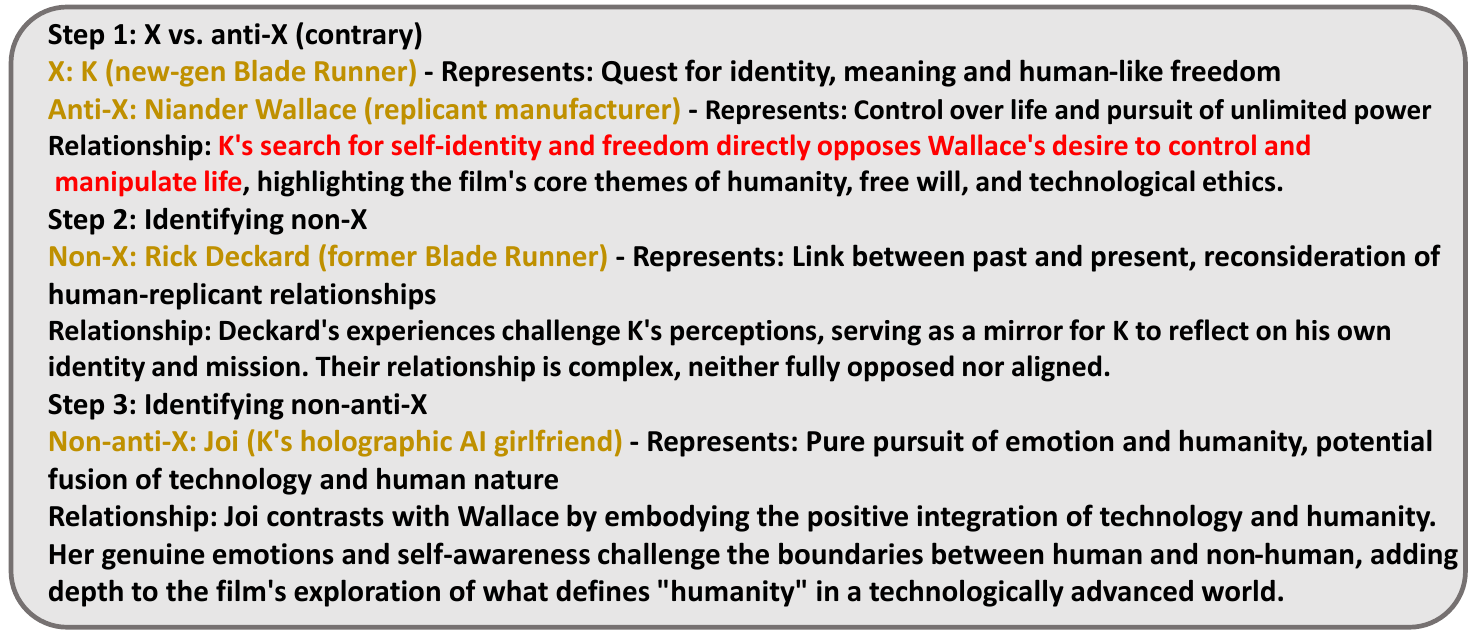}}
  \hfill
  \subcaptionbox{\textit{Les Misérables}}{\includegraphics[width=1\columnwidth]{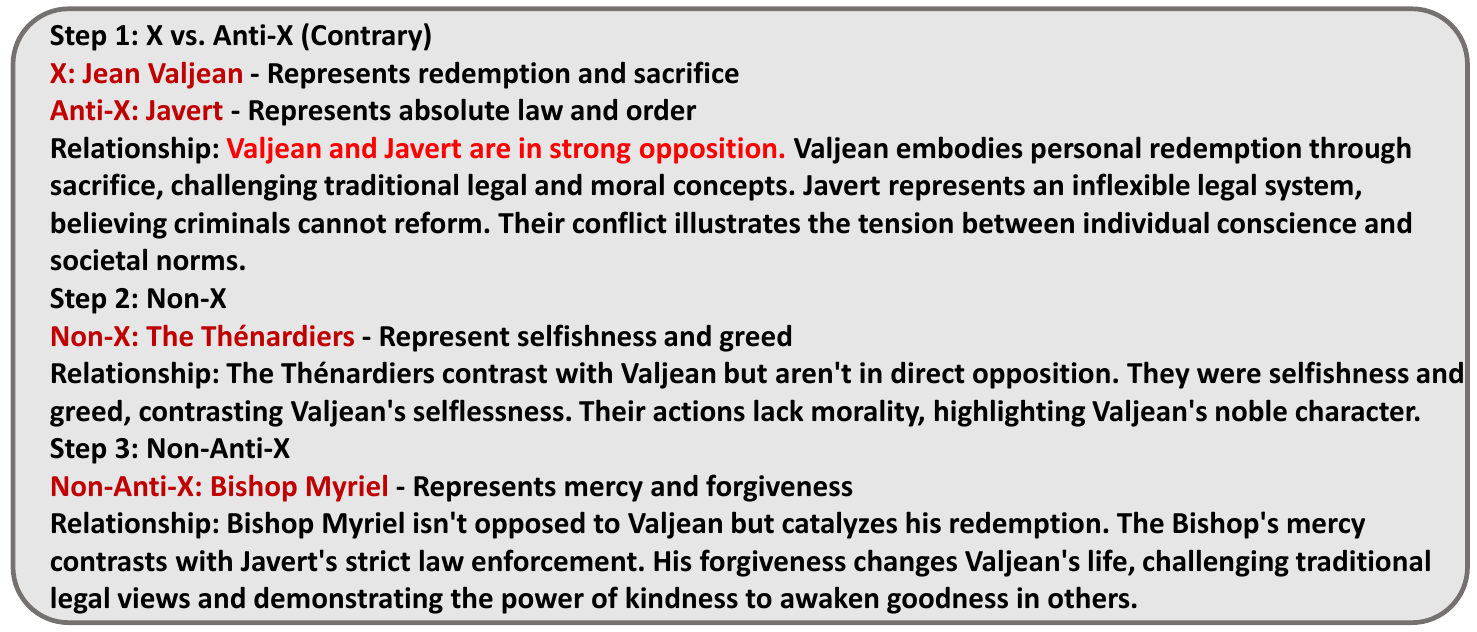}}
  \hfill
  \subcaptionbox{\textit{Anna Karenina}}{\includegraphics[width=1\columnwidth]{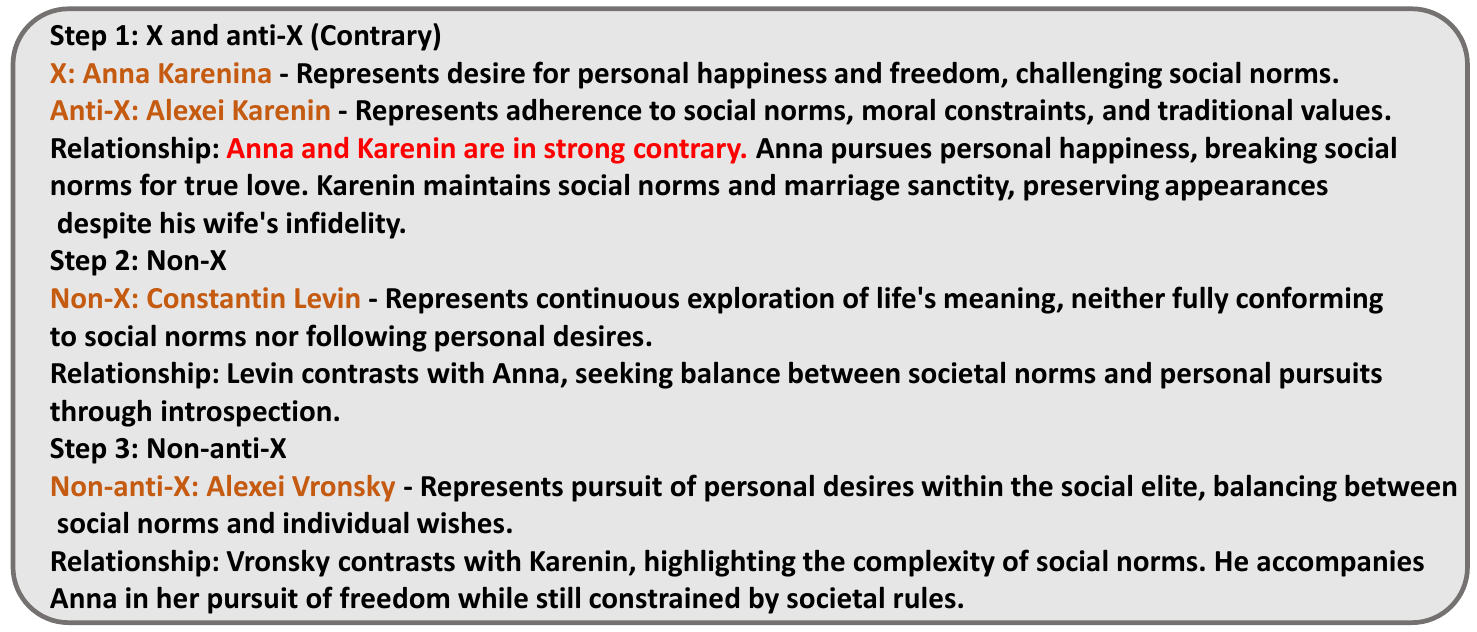}}
  \vspace{-5pt}
  \caption{Our proposed GLASS framework output: Greimas semiotic square analysis of (a) \textit{Blade Runner 2049}, (b) \textit{Les Misérables}, (c) \textit{Anna Karenina}.}
  \label{fig:GSSOutput}
\end{figure}

\section{Novel Semiotic Square Analysis}
We utilize GLASS (with Claude 3.5 Sonnet, Qwen2.5, DeepSeek-R1 and Kimi) for novel semiotic square analysis. We can demonstrate the versatility of the GLASS across diverse themes and narrative modes. Some of the results are illustrated in Figure \ref{fig:GSSOutput}. This analysis reveals \textit{Blade Runner 2049}'s complex thematic structure, exploring identity, human nature, technological ethics, and free will in a highly advanced future society. In \textit{Les Misérables}, exploring themes of law vs. morality, crime and punishment, and redemption vs. forgiveness. Hugo examines human nature's complexity and the balance between societal rules and individual conscience. In \textit{Anna Karenina}, this analysis reveals the core conflict: personal desire versus social norms in love and morality.

We applied the proposed framework to analyze another 39 classic or renowned narrative works, such as \textit{War and Peace}, \textit{Robinson Crusoe} and \textit{Hamlet}. Due to space limitations, detailed analyses generated by LLMs using the framework are provided here\footnote{\url{https://github.com/zengyfan/GSSDataset}}. To our knowledge, some of these works have not previously undergone structuralist semiotic square analysis, and our work contributes to literary studies in this regard.

\section{Conclusion}
In this paper, we propose a novel structuralist AI literary criticism framework, GLASS, which innovatively applies the Greimas semiotic square to enable LLMs to conduct more professional and in-depth analyses of narrative patterns and meaning structures in literary works. We constructed a Greimas semiotic square dataset through annotations by human literary scholars and incorporated it into our proposed framework using few-shot prompting and CoT. Then, we applied GLASS to conduct in-depth semiotic square literary criticism on more than 40 works. In order to quantitatively compare GLASS and human experts, we propose the quantitative metrics QEMG for GSS-based literary criticism using the LLM-as-a-judge paradigm. Our framework's results,  compared with human experts' criticism across multiple works and LLMs, show superior performance in various aspects. Additionally, we also specifically and intuitively demonstrate the excellent capabilities of GLASS on some classic works. In conclusion, our work provides new methods, perspectives, and inspirations from AI and computational linguistics for literary criticism, while assisting humans in literary studies.

\nocite{ChalnickBillman1988a}
\nocite{Feigenbaum1963a}
\nocite{Hill1983a}
\nocite{OhlssonLangley1985a}
\nocite{Matlock2001}
\nocite{NewellSimon1972a}
\nocite{ShragerLangley1990a}

\bibliographystyle{apacite}

\setlength{\bibleftmargin}{.125in}
\setlength{\bibindent}{-\bibleftmargin}

\bibliography{CogSci_Template}

\end{document}